\PassOptionsToPackage{dvipsnames}{xcolor}
\documentclass[sigconf, screen]{acmart}

\usepackage{booktabs} %

\usepackage{mathtools}

\usepackage{microtype}

\usepackage{etoolbox}
\robustify{\bfseries}
\robustify{\itshape}

\DeclarePairedDelimiterX{\dotp}[2]{\langle}{\rangle}{#1,#2}

\DeclareMathOperator{\simf}{\text{sim}}

\let\given\givenbase

\DeclarePairedDelimiterX{\infdivx}[2]{(}{)}{%
  #1\;\delimsize\|\;#2%
}
\newcommand{\kl}{\operatorname{KL}\infdivx}

\makeatletter
\patchcmd{\NAT@test}{\else \NAT@nm}{\else \NAT@nmfmt{\NAT@nm}}{}{}

\DeclareRobustCommand\citepos
{\begingroup
  \let\NAT@nmfmt\NAT@posfmt%
  \NAT@swafalse\let\NAT@ctype\z@\NAT@partrue
  \@ifstar{\NAT@fulltrue\NAT@citetp}{\NAT@fullfalse\NAT@citetp}}

\let\NAT@orig@nmfmt\NAT@nmfmt
\def\NAT@posfmt#1{\NAT@orig@nmfmt{#1's}}
\makeatother

\usepackage{tikz}
\usetikzlibrary{
  arrows.meta,
  calc,
  plotmarks,
  positioning,
}
\usepackage{pgfplots}
\usepackage{pgfplotstable}

\usepgfplotslibrary{
  colorbrewer,
}

\pgfplotsset{
  grid style=dashed,
  ymajorgrids=true,
  cycle list/Dark2-5,
  cycle multiindex list={
    Dark2-5\nextlist
    mark list
  },
}

\usepackage{siunitx}
\usepackage{booktabs}

\usepackage{xspace}
\makeatletter
\DeclareRobustCommand\onedot{\futurelet\@let@token\@onedot}
\def\@onedot{\ifx\@let@token.\else.\null\fi\xspace}

\def\eg{{e.g}\onedot} 
\def\ie{{i.e}\onedot} \def\Ie{{I.e}\onedot}
\def\cf{{cf}\onedot} 
 
\def\wrt{w.r.t\onedot}

\makeatother

\acmConference[SAC'22]{ACM SAC Conference}{April 25 –April 29, 2022}{Brno, Czech Republic}
\copyrightyear{2022}
\acmYear{2022}
\setcopyright{acmlicensed}\acmConference[SAC '22]{The 37th ACM/SIGAPP
  Symposium on Applied Computing}{April 25--29, 2022}{Virtual Event, }
\acmBooktitle{The 37th ACM/SIGAPP Symposium on Applied Computing (SAC'22), April 25--29, 2022, Virtual Event, }
\acmPrice{15.00}
\acmDOI{10.1145/3477314.3507267}
\acmISBN{978-1-4503-8713-2/22/04}

\acmArticle{4}
\acmPrice{15.00}

\usepackage{hyperref}
\hypersetup{
  colorlinks,
  linkcolor = BrickRed,
  citecolor = RoyalBlue,%
  urlcolor  = WildStrawberry,
}

\begin{document}
\title{Representation Learning via Consistent Assignment of Views to Clusters}

\author{Thalles Silva}
\affiliation{%
  \institution{University of Campinas}
  \streetaddress{Av. Albert Einstein, 1251}
  \city{Campinas}
  \state{SP}
  \postcode{13083-852}
}
\email{thalles.silva@students.ic.unicamp.br}

\author{Ad\'in Ram\'irez Rivera}
\orcid{0000-0002-4321-9075}
\affiliation{%
  \institution{Reykjavik University}
  \streetaddress{Menntavegi 1}
  \city{Reykjavík}
  \postcode{102}
}
\additionalaffiliation{%
  \institution{University of Oslo}
}
\additionalaffiliation{%
  \institution{University of Campinas}
}
\email{adinr@uio.no}

\renewcommand{\shortauthors}{Silva and Ram\'irez Rivera}

\begin{abstract}
	We introduce Consistent Assignment for Representation Learning (CARL), an unsupervised learning method to learn visual representations by combining ideas from self-supervised contrastive learning and deep clustering.
	By viewing contrastive learning from a clustering perspective, CARL learns unsupervised representations by learning a set of general prototypes that serve as energy anchors to enforce different views of a given image to be assigned to the same prototype.
	Unlike contemporary work on contrastive learning with deep clustering, CARL proposes to learn the set of general prototypes in an online fashion, using gradient descent without the necessity of using non-differentiable algorithms or K-Means to solve the cluster assignment problem.
	CARL surpasses its competitors in many representation learning benchmarks, including linear evaluation, semi-supervised learning, and transfer learning.
	Code at: \href{https://gitlab.com/mipl/carl/}{https://gitlab.com/mipl/carl/}.
\end{abstract}




\maketitle

\section{Introduction}

Unsupervised visual representation learning focuses on creating meaningful representations from data and inductive biases. 
Lately, methods based on Siamese neural networks~\cite{bromley1994signature} and contrastive loss functions~\cite{he2020momentum, chen2020simple} have significantly reduced the gap between supervised and unsupervised based representations. 
Indeed, for some downstream tasks, unsupervised-based representations already surpass their supervised counterparts~\cite{caron2020unsupervised}. 
In computer vision, approaches to unsupervised representation learning can be categorized into three groups: (1)~contrastive learning methods using instance discrimination, (2)~clustering-based methods, and (3)~a mixture of the two.

Recent state-of-the-art unsupervised representation learning rely on contrastive learning~\cite{tian2020makes, he2020momentum, chen2020simple, chen2020exploring, oord2018representation, chen2020big}.
These methods optimize an instance discrimination pretext task where each image and its transformations are treated as individual classes. 
They compare feature vectors of individual images intending to organize the feature space such that similar concepts are placed closer while moving different ones farther.

On the other hand, traditional clustering methods aim to learn the data manifold by comparing groups of features that share semantic structure based on a distance metric.
When combined with deep learning, clustering methods are often designed as two-step algorithms. 
First, the complete dataset is clustered, and then the meta clustering information, \eg, prototypes and pseudo-labels, are used as supervisory signals in a posterior optimization task~\cite{caron2018deep, caron2019unsupervised, asano2019self, yan2020clusterfit}.

Recent work has attempted to combine the benefits of contrastive learning and clustering~\cite{li2020prototypical, caron2020unsupervised}. 
In particular, Expectation-Maximization (EM) approaches alternate between finding the clusters and maximizing the mutual information (MI) between the embeddings and the cluster centroids~\cite{li2020prototypical}.
Inspired by them, our work merges the benefits of both approaches by bridging the gap between clustering and contrastive learning. 
On the one hand, we use unsupervised clustering dynamics to generate robust prototypes that organize the feature space.
On the other, we use contrastive learning to compare the distributions of the views' assignments \wrt the clusters. 
Our experiments show that we can learn unsupervised visual representations that outperform existing methods by mixing both approaches. 

We can think of contrastive learning as learning clustered representations at the image level. 
However, given the nature of the task, these clusters fail to capture semantic information from the heterogeneous unknown classes since the learned clusters only comprise representations from synthetic views of an image. 
Moreover, since contrastive learning methods handle different images as negatives in the training process, even if two distinct images share the same class information, their representations will be pushed farther apart from each other.
In the end, each image will have its own cluster structure.

We propose an alternative method to learn high-level features by clustering views based on consistent assignments. 
Unlike previous work that uses $K$-Nearest Neighbors~\cite{altman1992introduction} or $K$-Means~\cite{lloyd1982least} as priors to enforce (learn) a cluster mapping, our method learns the prototypes online.
While contrastive learning with instance discrimination~\cite{chen2020simple, he2020momentum} poses a classification pretext task at the image embedding level, we propose a pretext task that operates on the assignment of views to a set of clusters. 
Rather than directly maximizing similarities between image embeddings, we force the distribution of positive views' assignments to be consistent among a set of finite learnable prototypes.
If the number of prototypes equals the number of observations in the dataset, we would be forcing each cluster only to contain synthetic views of a given observation. 
This is equivalent to contrastive learning with instance discrimination.
However, if we set the number of prototypes to be smaller than the number of observations in the dataset, by the pigeonhole principle, the learned prototypes will not only cluster different views of an image together, but they will also aggregate representations of different images that are similar enough to be assigned to the same cluster. 

Regarding our contributions, (i)~we propose a learning framework that leverages current contrastive learning methods with clustering-based algorithms to improve the learned representations. 
Unlike contemporary work, our method proposes to learn the clusters' assignments in an online fashion using gradient descent with no need for pre-clustering steps \eg, $K$-Means or non-differentiable methods to solve the clustering assignment problem, such as the Sinkhorn-Knopp algorithm~\cite{cuturi2013sinkhorn}.
(ii)~We contrast high-level structures (the distributions of the views over the cluster assignments) instead of low-level ones (such as the representations). 
Reasoning in this high-level space gives the representations more robustness that translates to better performance in downstream tasks. 
And, (iii)~unlike contrastive learning methods that can be viewed as learning clusters containing only synthetic transformations of a given image, our learned prototypes do not need to hold the semantics of the data but rather become energy anchors that self-organize the space to learn better representations.
Moreover, the proposed loss function does not require a more extensive set of negative representations, which avoids the common problem of treating representations of the same class as negatives.

\section{Related Work}

Self-supervised learning relates to the idea of extracting supervisory signals from the data. 
Methods including relative patch prediction~\cite{doersch2015unsupervised}, jigsaw puzzle~\cite{noroozi2016unsupervised}, rotation prediction~\cite{gidaris2018unsupervised}, and colorization~\cite{zhang2016colorful} propose manually crafted pretext tasks that, when optimized, can make a deep neural network learn useful representations without the need of human-annotated datasets. 
Most of these methods work with the same principle. 
They corrupt the input with stochastic random transformations and challenge the network to predict some property of the corrupted input.

One such pretext task is instance discrimination~\cite{dosovitskiy2015discriminative, wu2018unsupervised}. 
It describes a classification task in which each image is treated as a unique class and, therefore, stochastic transformations of the same image, called views, should belong to the same class. 
\citet{dosovitskiy2015discriminative} proposed to optimize this task by learning a linear classifier where the number of output classes matches the number of observations in the dataset. 
Following, \citet{wu2018unsupervised} proposed to use a Noise-Contrastive Estimation~(NCE) approximation of the non-parametric softmax classifier that could scale to large datasets. 

Currently, contrastive learning~\cite{hadsell2006dimensionality} methods rely on an NCE based loss function called InfoNCE~\cite{oord2018representation, tian2019contrastive}. 
Recent work describes optimizing the InfoNCE loss through the lens of maximizing the MI between representations of the same image~\cite{hjelm2018learning, henaff2020data, bachman2019learning}. 
InfoNCE characterizes an $M+1$ classification task where a pair of positive examples needs to be identified among the set of $M$ negatives.
In practice, the success of InfoNCE requires a large number of negative examples.
Nonetheless, since negatives are usually randomly sampled from the dataset, it often leads to a false-negative problem where representations from images of the same class are treated as negatives~\cite{saunshi2019theoretical}. 

\citet{he2020momentum} and \citet{chen2020improved} presented MoCo, a contrastive learning framework that employs an additional momentum encoder to provide consistent instance representations for the InfoNCE loss. 
\citet{chen2020simple} presented SimCLR, a Siamese-based~\cite{bromley1994signature} contrastive learning method trained with InfoNCE that relies on large batch sizes to draw a high number of negative samples. 
BYOL~\cite{grill2020bootstrap} proposes a framework that does not require negative samples and learns visual representations by approximating augmented views of the same data point using an $\ell_2$ loss in the latent space. 
Unlike contrastive learning with instance discrimination, we seek to learn prototype vectors that act as anchors in the embedding space. 
We use these anchors as energy beacons for the images. 
Our goal is to use the energy distributions induced by the similarity of the images \wrt the prototypes to find representations that share similarities in the feature space. 

Recent work proposed clustering-based methods for deep unsupervised representation learning~\cite{asano2019self, yan2020clusterfit, caron2019unsupervised, li2020prototypical, caron2020unsupervised}. 
DeepCluster~\cite{caron2018deep} learns representations by predicting cluster assignments. 
Once per epoch, the algorithm clusters the entire dataset using $K$-Means and uses the pseudo-labels as supervised signals in an additional classification task. 
One of the limitations of this approach is that the classification layer needs to be reinitialized once per clusterization. 
DeeperCluster~\cite{caron2019unsupervised} builds on top of \citepos{caron2018deep} work and presents an algorithm to combine hierarchical clustering with unsupervised feature learning using the rotation prediction pretext task~\cite{gidaris2018unsupervised}. 
Similarly, Prototypical Contrastive Learning (PCL)~\cite{li2020prototypical}, formulates a self-supervised visual representation learning framework as an Expectation-Maximization algorithm. 

Our method utilizes a conceptually distinct methodology. 
Unlike \citepos{caron2018deep} and \citepos{li2020prototypical} works, our method does not require a pre-clusterization step of the entire corpus, which vastly reduces memory and computing power requirements. 
Moreover, since we do not use $K$-Means clustering as a proxy to learn an additional task, we do not need to reinitialize any layers during optimization, nor is our method susceptible to limitations and assumptions implied by the $K$-Means algorithm. 
Instead, we propose to learn the prototypes end-to-end by consistently enforcing different views of the same image to be assigned to the same prototypes. 
Lastly, our method does not rely on handcrafted pretext tasks. 

Similar to our work, \citet{caron2020unsupervised} proposes an online clustering method to learn visual representations by contrasting cluster assignments on a second set of latent variables. 
To avoid collapsing modes where all observations are assigned to a few classes, they use the Sinkhorn-Knopp algorithm~\cite{cuturi2013sinkhorn} to solve the cluster assignment problem over the latent variables to guide the encoder during learning. 
Unlike \citepos{caron2020unsupervised} work, our method learns cluster assignments end-to-end via gradient descent and avoids trivial solutions by enforcing a regularization term over the cluster assignments of views in a given batch. 

\citet{van2020learning} presented a two-step algorithm for unsupervised classification and proposed the SCAN-Loss (Semantic Clustering by Adopting Nearest neighbors) as part of the learning pipeline. 
The idea is to decouple feature learning and clustering by learning two encoders in distinct optimization stages. 
The algorithm extracts a set of nearest neighbors from each observation and uses them as priors to learn a second network for semantic clustering. 
The nearest neighbors are mined from representations of an encoder trained using a self-supervised pretext task. 
Our implementation builds on top of the SCAN-Loss, but unlike \citet{van2020learning}, we employ a Siamese network architecture to learn representations via cluster assignments, end-to-end, without the necessity of optimizing for a second self-supervised task or mining of nearest neighbors. 
Our objective is to compare consistent assignments to a set of clusters for each view instead of comparing the underlying representations nor their true clusters. 
By contrasting high-level representations of the data, we obtain reliable data representations that perform well in downstream tasks.

\section{Proposed Method}
\begin{figure*}
\centering
\begin{tikzpicture}[
  border/.style={
    inner sep=0pt,
    outer sep=0pt,
    clip,
    rounded corners,
  },
  connector/.style={
    draw=gray,
    shorten <=3pt,
    shorten >=3pt,
    rounded corners,
  },
  proto/.style={
    text=BrickRed,
  },
  proc/.style={
    rectangle,
    rounded corners,
    minimum width=2.2cm,
    minimum height=1cm,
  },
  proc/.default=white,
  solid/.style={
    draw=#1,
    fill=#1,
  },
  node/.style={
    draw,
    circle, 
    minimum width=1.cm,
  },
  loss/.style={
    node,
    dashed,
  },
  edge/.style={
    >=Latex,
    ->,
    shorten <=3pt,
    shorten >=3pt,
    rounded corners,
  },
  declare function={
    gaussian(\x,\mu,\sigma) = 1/(\sigma*sqrt(2*pi))*exp(-((\x-\mu)^2)/(2*\sigma^2));
    gamma(\z)= (2.506628274631*sqrt(1/\z) + 0.20888568*(1/\z)^(1.5) + 0.00870357*(1/\z)^(2.5) - (174.2106599*(1/\z)^(3.5))/25920 - (715.6423511*(1/\z)^(4.5))/1244160)*exp((-ln(1/\z)-1)*\z);
    beta(\x,\alpha,\beta) = gamma(\alpha+\beta)*\x^(\alpha-1)*(1-\x)^(\beta-1) / (gamma(\alpha)*gamma(\beta));
    gammapdf(\x,\k,\theta) = \x^(\k-1)*exp(-\x/\theta) / (\theta^\k*gamma(\k));
  },
] 
  \newlength{\clusterall}
  \newcommand{\cluster}[3][]{
  \pgfkeys{
    /tikz/.cd,
    node sep/.store in=\clustersep,
    node sep/.default=.5cm,
    mark/.store in=\mark,
    mark/.default=*,
    cluster node hidden/.style={},
    cluster node/.style={cluster node hidden/.append style={##1}},
    mark,
    node sep,
    #1
  }
    \begin{scope}
      \coordinate (#2) at #3;
      \pgfmathsetlengthmacro{\clusterall}{1.2*\clustersep}
      \draw[cluster node hidden, fill, fill opacity=.1, draw opacity=0.1] (#2) circle (\clusterall);
      \node[cluster node hidden] at (#2) {\pgfuseplotmark{\mark}};
      \foreach \angle [count=\i] in {0,60,...,360}
        \node[cluster node hidden] (#2-\i) at ($(#2)+(\angle:\clustersep)$) {\pgfuseplotmark{\mark}};
    \end{scope}
  }
  \cluster[cluster node={text=Dark2-A,Dark2-A}, node sep=1cm]{a}{(0,0)}
  \cluster[cluster node={text=Dark2-B,Dark2-B}, node sep=.75cm]{b}{(2.5,0)}
  \cluster[cluster node={text=Dark2-C,Dark2-C}, node sep=0.5cm]{c}{(1.25,2)}
  
  \node[left=1 of a-4, border] (i2) {\includegraphics[width=1cm]{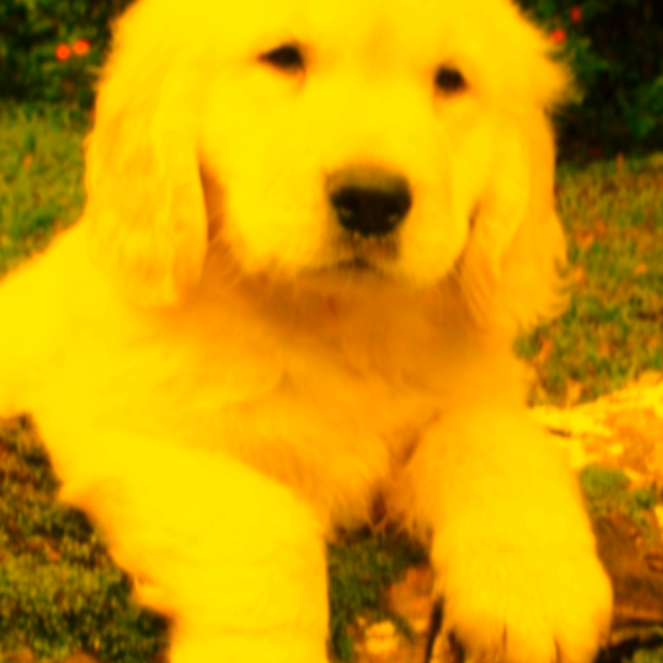}};
  \node[above=2pt of i2, border] (i1) {\includegraphics[width=1cm]{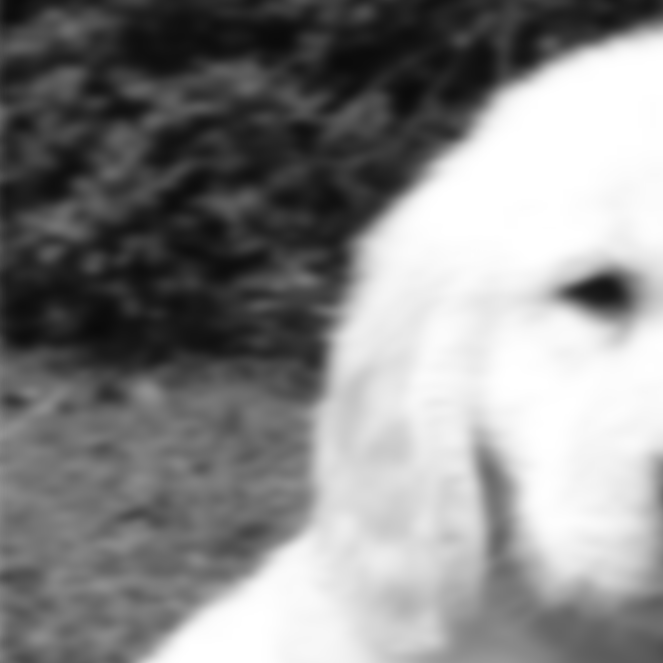}};
  \node[above=2pt of i1, border] (i) {\includegraphics[width=1cm]{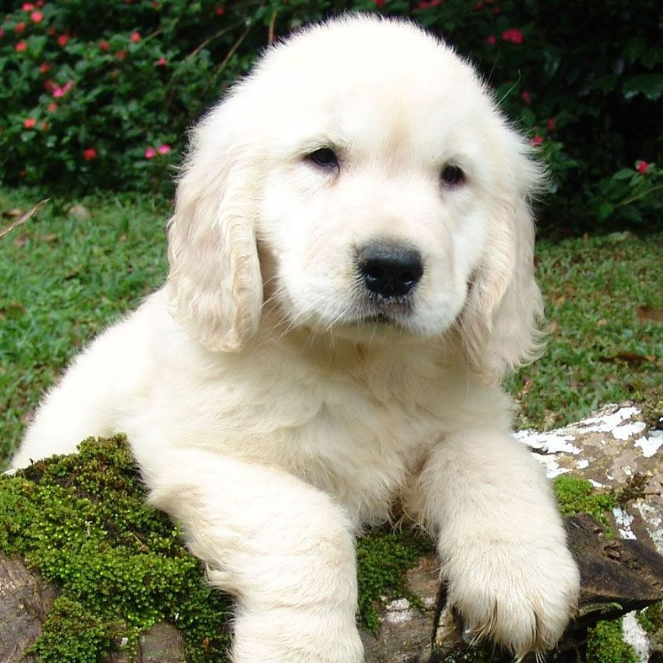}};
  
  \draw[connector] (i) -| (a);
  \draw[connector] (i1) -| (a-3);
  \draw[connector] (i2) -- (a-4);
  
  \node[left=1pt of i] {$x_i$};
  \node[left=1pt of i1] {$v_i^a$};
  \node[left=1pt of i2] {$v_i^+$};
  
  \node[below=1pt of a-3] {$z_i^a$};
  \node[below=1pt of a-4] {$z_i^+$};
  
  \node[proto] (p1) at (-0.25,1.65) {\pgfuseplotmark{triangle*}};
  \node[proto] (p2) at (2.75,.95) {\pgfuseplotmark{triangle*}};
  \node[proto] (p3) at (1.5,-.5) {\pgfuseplotmark{triangle*}};
  \node[proto] (p4) at (2.2,2) {\pgfuseplotmark{triangle*}};
  
  \node[right=2pt of p4] {Prototypes};
  
  \foreach \i in {1,2,3,4}{
    \draw[connector, dashed, WildStrawberry] (a-3) -- (p\i);
    \draw[connector, dashed, BurntOrange] (a-4) -- (p\i);
    \draw[connector, dashed, BlueViolet] (c-5) -- (p\i);
  }
  
  \node[right=1 of a-6 -| b-1, anchor=south west, inner sep=0pt] (dist) {%
    \resizebox{4cm}{2.5cm}{%
      \begin{tikzpicture}[]%
        \begin{axis}[
          samples=100,
          y axis line style={draw=none},
          x axis line style={line width=1pt},
          grid=none,
          tick style={draw=none},
          axis x line*=bottom,
          ticks=none,
          enlarge y limits=false,
          line width=3pt,
          ]%
          \addplot [smooth, domain=0:1, draw=none, fill=WildStrawberry, fill opacity=.5] {beta(x,3,7)};%
          \addplot [smooth, domain=0:1, draw=none, fill=BurntOrange, fill opacity=.5] {beta(x,1.5,5)};%
          \addplot [smooth, domain=0:1, draw=none, fill=BlueViolet, fill opacity=.5] {beta(x,5,2)};%
        \end{axis}%
      \end{tikzpicture}%
  }};
  
  \node[above=5pt of i] (dl) {Data};
  \node at (dl -| {$(a)!.5!(b)$}) {Representations};
  \node at (dl -| dist) {Assignments};
  
\end{tikzpicture}
\caption{The data is first augmented and then encoded into their representations.  We learn a set of \textcolor{BrickRed}{prototypes} that serve as anchors to compute a distribution of assignments.  For a \textcolor{Dark2-A}{positive (augmented) pair of points} and for the rest of \textcolor{Dark2-C}{negatives} we compute their agreement to the prototypes.  We collect these comparisons as distributions that we later compare.  We expect that the positive examples will have similar distributions over the assignments (\cf orange and pink distributions), while the negatives will not (\cf purple distribution).}
\label{fig:teaser}
\end{figure*}
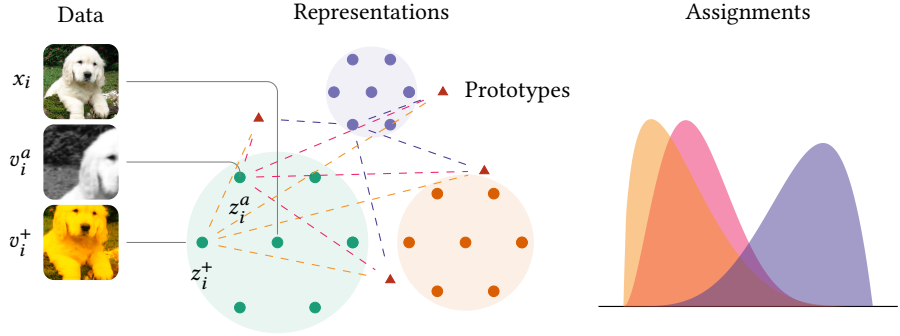

Our proposal, Consistent Assignment for Representation Learning (CARL), re-frames Contrastive Learning through a clustering perspective to learn robust representations (Section~\ref{sec:cl-clust}).
CARL builds distributions of the similarities between the prototypes and the image's views (Section~\ref{sec:carl}). 
To avoid collapsing the representation to a single prototype, we impose an uninformative prior to CARL's prototypes (Section~\ref{sec:entropy}). 
CARL jointly optimizes similarity to the learned prototypes and the uninformative prior (with a decay schedule for learning). 
Fig.~\ref{fig:teaser} illustrates the overall idea.

\subsection{Contrastive Learning from a Clustering Perspective}
\label{sec:cl-clust}

Let $\mathcal{X} = \{x_1, x_2, \dots, x_N\}$ be a dataset containing $N$ unlabeled images. 
And, let a view~$v_i = T(x_i)$ of an observation~$x_i$ be the application of a stochastic function $T$ that is designed to change the content of $x_i$ subjected to preserving the task-relevant information encoded in it. 
In practice, we can create as many views as needed by applying the stochastic function~$T$. 
Recent contrastive learning methods learn visual embeddings by solving an instance discrimination pretext task that is usually optimized using the InfoNCE loss~\cite{oord2018representation} defined as
\begin{equation}
\label{eqn:infonce-loss}
\mathcal{L}_\text{InfoNCE} = - \log \frac{\exp\left(\simf \left(z^a_i, z^+_i \right)/\tau \right)}{\sum_{j}^M \exp \left(\simf \left(z^a_i, z_j \right)/\tau \right)},
\end{equation}
where $z^a_i$ and  $z^+_i$ are anchor and positive representations taken from an encoder function $f$ such that $z_i^{(\cdot)} = f\big(v_i^{(\cdot)}\big)$, $\tau$ is the temperature parameter, and $\simf(\cdot, \cdot)$ is a similarity function, \eg, the cosine similarity. 
From a clustering perspective, the InfoNCE loss~\eqref{eqn:infonce-loss} is minimized when all possible variants $\big\{v_i^j\big\}_j$ of an image $x_i$ are clustered into the same prototype while representations from within a cluster are far apart from representations of other clusters.

We propose an approach where, instead of comparing against other instances~\cite{he2020momentum} or prototypes of the classes~\cite{li2020prototypical}, we learn a set of $K$~general prototypes $\mathcal{C} = \{c_1, c_2, \dots, c_K\}$, $K \ll N$, against which we compare the views to determine their similarity. 
Instead of maximizing a similarity function between positive embeddings of different views, as contrastive learning methods do, we maximize the similarity between assignment vectors of positive views to our general prototypes to promote consistency and confidence when assigning views to clusters.
\Ie, views must agree with high confidence in their cluster assignments. 
Our method learns the prototype matrix~$\mathcal{C}$ online using gradient descent, and it does not require any pre-clusterization step as is typically the case for clustering-based representation learning algorithms~\cite{li2020prototypical, caron2018deep}.
Moreover, since the training process is unsupervised, to avoid trivial solutions, where all images are assigned to only a handful of prototypes, we enforce a non-informative prior over the class distribution of the views to guide the learning process.

\subsection{Consistent Assignment for Representation Learning}
\label{sec:carl}

As with previous methods, we treat augmented versions of a given image as views and use them as positive examples for optimization.
Our objective is to transform two positive samples into a distribution of their likelihood to belong to a set of $K$~clusters.
To do so, we encode each view through an encoder function $z_i = f(v_i) \in \mathbb{R}^d$.
The encoder~$f$ comprises a backbone convolutional neural network, such as a ResNet~\cite{he2016deep}, followed by a non-linear multilayer perceptron (MLP) head. 

Our objective is not to cluster the data in an unsupervised manner but rather to learn a set of prototypes that will serve as anchors to differentiate views. 
We hypothesize that similar views should have similar assignments \wrt the prototypes. 
Hence, to convert the representations into these assignments, we first compare the representation~$z_i$ against all the prototypes to obtain an energy distribution 
\begin{equation}
  q_i[j] = \dotp{z_i}{c_j},
\end{equation}
where $q_i[j]$ is the $j$-th element of the energy distribution~$q_i$ for the $i$-th view.  
We learn the set of prototypes through a linear layer. 
To get the distribution of a view over all prototypes, we normalize the energy using the softmax function and obtain the posterior probability distribution, \ie, the probability of assigning the view $v_i$ to the cluster $k$.
Hence, our normalized probability for the $k$-th class given our view $v_i$ is
\begin{equation}
\label{eqn:softmax}
p_i[k] = P(i \text{ assigned to } k \given v_i) = \frac{\exp(q_i[k])}{\sum_{j=1}^K \exp(q_i[j])},
\end{equation}
where $q_i[j]$ denotes the $j$-th element of the $i$-th un-normalized vector output of the classifier for the respective view.

As mentioned before, our objective is to contrast the distributions of two views' likelihood \wrt the clusters.
In CARL, the encoder and assigner operate in a Siamese setup where a pair of views from an image is independently transformed in its corresponding distribution.  
To ensure the similarity between the views, we optimize the views' distributions $p_i^a$ and~$p_i^+$ over the clusters in $\mathcal{C}$, so that the two distributions are consistent with one another. 
In other words, by learning a consistent assignment of views over the clusters, a given prototype will be invariant to augmented versions of an input image. 
Moreover, because the number of prototypes is smaller than the number of observations in the dataset, the clusters will contain different observations that share similarities in the embedding space.

We compute the similarity between the views' distributions as their dot product
\begin{equation}
\label{eqn:consistency-loss}
\mathcal{L}_c = - \frac{1}{B} \sum^B_i \log \dotp{p^a_i}{p^+_i},
\end{equation}
where $B$ is the size of a minibatch over which we are aggregating the samples.
In the ideal case of two one-hot vectors signaling the same perfect assignment, the dot product above yields its maximum value of one, and the negative $\log$ is minimized. 

\subsection{Preventing Trivial Solutions}
\label{sec:entropy}

Only forcing different views, $v^a_i$ and~$v^+_i$, to have the same cluster assignment using our consistency loss~\eqref{eqn:consistency-loss} leads to finding a trivial solution where all representations~$z_i$ end up assigned to the same cluster (or just to a handful of them).  
To prevent such triviality, we force the distribution over the classes, $P$, to be uninformative by minimizing the Kullback-Leibler divergence \wrt a uniform distribution, $U$.
Our regularization is
\begin{equation}
\label{eqn:entropy-reg}
\mathcal{L}_{\text{KL}} = \kl{P}{U} = \log(K) + \sum_{c \in C} \hat{p}^c \log\left ( \hat{p}^c\right ),
\end{equation}
where $\hat{p}^c$ is the expected distribution over a minibatch of size~$B$,
\begin{equation}
\label{eqn:entropy-sum}
\hat{p}^c = \frac{1}{B} \sum^B_i p^c_i.
\end{equation}
In other words, we maximize the Shannon entropy of the average distribution of the predictions.
We can interpret the KL-divergence~\eqref{eqn:entropy-reg} as regularizing the encoder $f$ to encourage the approximate posterior~\eqref{eqn:softmax} to be closer to the uniform distribution.

Minimizing the KL-divergence~\eqref{eqn:entropy-reg} will force the predictions for a view~$v_i$ to be spread across all clusters.
Since we do not know the underlying class distribution in advance, the KL-divergence~\eqref{eqn:entropy-reg} acts as a prior where we assume that the observations~$\mathcal{X}$ are uniformly assigned among all $K$~prototypes.

By combining the consistency assignment loss~\eqref{eqn:consistency-loss} with the KL-divergence regularization~\eqref{eqn:entropy-reg} we obtain our final learning objective
\begin{equation}
\label{eqn:final-loss}
\mathcal{L} = \mathcal{L}_c + \lambda_e \mathcal{L}_{\text{KL}},
\end{equation}
where $\lambda_e$ is an epoch-dependent function that returns a scalar that prevents mode collapse at the beginning of training. 
We observed that training is very susceptible to such collapsing to a single assignment if not regularized. 
However, in practice, we noticed that keeping a large fixed value of $\lambda_e$ during training also prevents the encoder from learning more complex representations. 
Thus, we recommend a function~$\lambda_e$ that decreases as training progresses.
In theory, any decay schedule, such as an exponential or cosine, could be used. 
We propose a linear decay schedule 
\begin{equation}
\lambda_e (a, b) = \begin{cases}
b - \frac{b-a}{E} e & \text{if $e \leq E$},\\
a & \text{otherwise},
\end{cases}
\label{eqn:linear-decay}
\end{equation}
where $b$ and $a$ denote the start and ending values of the decay, $E$ represents the number of epochs in which the decay will happen, and $e$ is the epoch counter. 

We noticed that mode collapse can happen in two scenarios: (1)~if the regularizer is not added to the final loss, which is equivalent to setting the KL weight penalty $\lambda_e(0,0)$, or (2)~if the KL weight penalty $\lambda_e$ is set to a value below a certain threshold, \cf Fig.~\ref{fig:ablation-lambda}.
In both situations, CARL finds suboptimal solutions and learns bad representations.
In practice, we found that slowly decreasing the KL weight penalty throughout training can work twofold since it increases the quality of the representations and prevents such trivialities during optimization. 
Refer to Section~\ref{subsec:kl-weight-penalty-ablation} and Fig.~\ref{fig:ablation-lambda} for an in-depth analysis. 

\section{Hyper-Parameter Exploration}
\label{sec:ablations}

In this section, we evaluate the effects of the primary hyperparameters of our method. 
For exploring hyper-parameters, we learn representations using a ResNet-18 backbone trained for \num{150} epochs, and the KL weight penalty $\lambda_e$ is linearly decayed over the first $E=100$ epochs.
To evaluate the multiple experimental setups, we train linear classifiers on top of the encoder's frozen features following the linear evaluation protocol proposed by \citet{he2020momentum} and report average Top-$1$ accuracy over three independent runs.

\subsection{Does Decreasing the KL Weight Penalty Improves Representation Learning?}
\label{subsec:kl-weight-penalty-ablation}

The hyperparameter $\lambda_e$ controls the contribution of the KL regularization~\eqref{eqn:entropy-reg} to the consistency loss~\eqref{eqn:consistency-loss}. 
Especially at the beginning of training, a higher contribution for the KL term avoids mode collapsing, where the network optimizes the consistency loss~\eqref{eqn:consistency-loss} by assigning all observations to the same prototype. 
\citet{van2020learning} make similar claims for the entropy regularization in their SCAN-loss~\cite{van2020learning}, and suggest a high (constant) value for the scalar hyperparameter $\lambda_e$ to avoid such trivialities.

\begin{figure}
\centering
\begin{tikzpicture}
\pgfplotstableread{
x	col1	col2	col3	col4	col5
50	71.84533333	70.92466667	31.6125 32.0750	69.587
100	75.77033333	74.658 30.5125	13.2250 73.850
150	78.287	76.51233333 30.3935	14.350	76.854
200	79.88733333 77.54566667 29.1185	13.4875 77.087

	}\data
\begin{axis}[
  small,
  xlabel={Training epochs},
  ylabel={Top-$1$ accuracy},
  legend style={
    at={(0.95,0.58)},
    anchor=east,
    font=\footnotesize,
  },
  legend cell align=left,
]
\addplot table [x=x, y=col5] from \data;
\addlegendentry{$\lambda_e(3,3)$}
\addplot table [x=x, y=col2] from \data;
\addlegendentry{$\lambda_e(2,2)$}
\addplot table [x=x, y=col3] from \data;
\addlegendentry{$\lambda_e(1,1)$}
\addplot table [x=x, y=col1] from \data;
\addlegendentry{$\lambda_e(1,2)$}
\addplot table [x=x, y=col4] from \data;
\addlegendentry{$\lambda_e(0,0)$}
\end{axis}
\end{tikzpicture}
\caption{%
  Effect of the uninformative prior's scheduling~$\lambda_e$ on the overall performance in STL-10. 
  Notice that a linear decay scheduling outperforms its constant counterparts. 
  Two situations may result in non-optimal solutions: (1)~a lower value of~$\lambda_e$, and (2)~not having the KL regularizer in the final loss. 
}
\label{fig:ablation-lambda}
\end{figure}
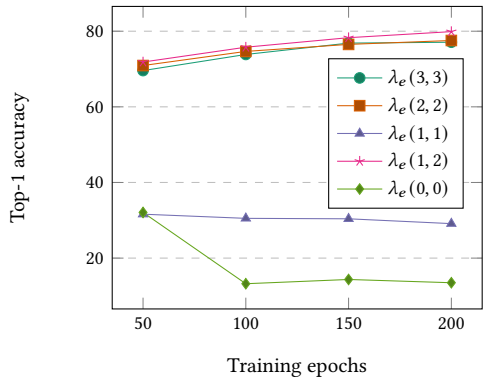

We hypothesize that keeping a high value of $\lambda_e$ throughout training prevents the network from learning complex features.
To verify this hypothesis, we trained CARL on the STL-10~\cite{coates2011analysis} unsupervised dataset for \num{200} epochs.
We measure the performance by training a linear classifier on top of the frozen features of the ResNet-18 backbone.
We linearly decay the magnitude of the $\lambda_e$ hyperparameter, following~\eqref{eqn:linear-decay}, from $b=2$ to $a=1$ over the first $E=100$ epochs instead of keeping it constant for one of the experiments. 
As shown in Fig.~\ref{fig:ablation-lambda}, we observe that the quality of the representations learned by CARL benefits from decreasing the contribution of the KL regularization.
However, if the $\lambda_e$ is decreased below a certain threshold, mode collapse may happen. 

\subsection{Does the Number of General Prototypes Influence the Quality of the Representations?}

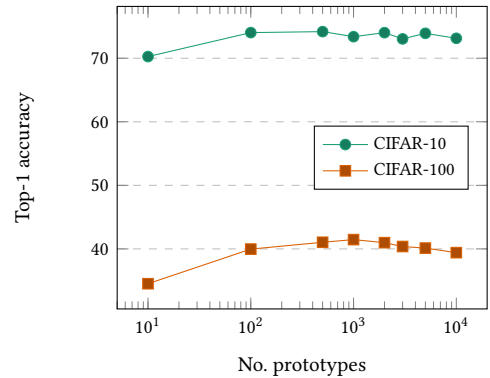
\begin{figure}
\centering
\begin{tikzpicture}
\pgfplotstableread{
x	CIFAR-10	CIFAR-100
10	70.25933333	34.52233333
100	74.029	39.9693
500	74.18333333	41.04266667
1000	73.38933333	41.463
2000	74.01966667	40.979
3000	73.04233333	40.376
5000	73.91633333	40.126
10000	73.13266667	39.40933333
}\data
\begin{axis}[
  small,
  xlabel={No.\ prototypes},
  ylabel={Top-$1$ accuracy},
  xmode=log,
  legend style={
    at={(0.95,0.5)},
    anchor=east,
    font=\footnotesize,
  },
  legend cell align=left,
]
\addplot table [x=x, y=CIFAR-10] from \data;
\addlegendentry{CIFAR-10}
\addplot table [x=x, y=CIFAR-100] from \data;
\addlegendentry{CIFAR-100}
\end{axis}
\end{tikzpicture}
\caption{%
  Effect of learning a different number of prototypes on the quality of the representations.
  Empirical tests suggest an inverse U-shape curve where the optimal number of prototypes lies near one order of magnitude \wrt the actual number of classes of the dataset.
}
\label{fig:prototypes-ablation}
\end{figure}

To evaluate the effect of learning a different number of prototypes, we trained CARL with ResNet-18 backbones on the CIFAR-10/100 datasets~\cite{krizhevsky2009learning} for $150$ epochs. 
For all experiments, the KL weight term $\lambda_e$ starts as $b=4$ and is linearly decreased to $a=1.5$ over the first $E=100$ epochs. 
Fig.~\ref{fig:prototypes-ablation} shows that over-clustering benefits the quality of the learned representations and that the optimal number of general prototypes depends on the number of actual classes of the dataset.
The experiments suggest an inverse U-shape for both datasets, and the optimal number of general prototypes lies near an order of magnitude \wrt the actual number of classes.

\subsection{Does the Batch Size Improves the Representations Learned by CARL?}

We regularize CARL using a non-informative prior~\eqref{eqn:entropy-reg} applied over the input batch. 
This uniform prior prevents the views' assignment from collapsing to a single prototype. 
We work over the expected distribution of assignments instead of each individual one. 
This expectation raises the question of how large a sample (batch size) should be to model different numbers of general prototypes.

Table~\ref{tab:ablations-batch-size} shows the effect of training CARL with different batch sizes and their relationship with the number of general prototypes.
For this experiment, we trained CARL on CIFAR-100 for \num{150} epochs. 
Note that to compensate for the reduced number of iterations (as we increase the batch size), we scale the learning rate following the recipe proposed by \citet{goyal2018accurate}.

With smaller batches, we can see that the performance of the representations learned by CARL degrades as the number of prototypes increases.
On the other hand, large batch sizes tend to retain high performance even when we increase the number of prototypes to very large numbers.
Indeed, the gap in performance due to the batch size can reach nearly \num{6} points of average accuracy when optimizing a downstream task.
We hypothesize that this is a direct effect of the sample size used to estimate the entropy in~\eqref{eqn:entropy-reg}.
In other words, as the distribution increases (with a higher number of general prototypes), the KL regularizer requires a larger batch size to better estimate the entropy and force it to be high~\eqref{eqn:entropy-reg}.

\begin{table}
  \footnotesize
  \sisetup{
    round-mode = places,
    round-precision=2,
    table-format=2.2,
    detect-all,
    table-text-alignment=center,
  }
  \begin{center}
  \caption{The effect of training CARL (on CIFAR-100) with different batch sizes and general prototypes.}
  \label{tab:ablations-batch-size}
    \newlength{\colsep}\setlength{\colsep}{.5em}
    \begin{tabular}{
        @{ }lSSSSSSS@{ }
      }
      \toprule
      & \multicolumn{7}{c}{Number of prototypes}\\
      \cmidrule{2-8}
      Batch & {10} & {100} & {500} & {1000} & {3000} & {5000} & {10000}\\

      \midrule
      64   & \textbf{35.34} & 38.719 & 38.599 & 37.639 & 35.979  & 35.75 & 34.12 \\
      128  & 35.25 & 39.43 & 39.5 & 38.969 & 38.379 & 38.59 & 36.329 \\
      256  & 34.459 & 40.309 & 40.509 & 40.879 & 39.979 & 38.829 & 39.18 \\
      512  & 35.009 & 40.909 & \textbf{42.559} & 41.229 & 41.059 & 40.61 & 38.649 \\
      1024 & 34.669 & \textbf{41.259} & 41.439 & \textbf{42.049} & \textbf{41.479} & \textbf{41.548} & \textbf{40.689}\\
      \bottomrule
    \end{tabular}
  \end{center}
\end{table}

\section{Unsupervised Feature Evaluation}
\label{sec:feature-eval}

We evaluate CARL's representations extracted from a ResNet-18 encoder and compare its performance with different state-of-the-art methods using the linear evaluation protocol~\cite{he2020momentum}, cross-domain transfer, semi-supervised learning, and fine-tuning. 

\subsubsection{Linear Evaluation}

Models were trained for \num{200} epochs, closely following their original hyper-parameters. 
Refer to Appendix~\ref{sec:appendix-A} for more details.
Table~\ref{tab:results-small-datasets} compare results with previous approaches on CIFAR-10/100~\cite{krizhevsky2009learning}, STL-10~\cite{coates2011analysis}, and on a downsampled version (DS) of the ImageNet dataset~\cite{chrabaszcz2017downsampled}.
We compare the representations learned by CARL against two methods based on instance discrimination with contrastive learning, \ie, SimCLR and MoCo v2, against two clustering-based methods, \ie, PCL and SwAV, and for completeness, we also include BYOL, which does not rely on negatives during optimization. 

CARL outperforms all major competitors in most setups and only stays behind on early training cases (100 epochs) for the STL-10 and DS ImageNet datasets.
Unlike SimCLR and MoCo~v2, our method does not use negative examples, nor does it require a momentum-based target encoder to prevent collapsing~\cite{grill2020bootstrap}.

We adapted PCL and SwAV to use a ResNet-18 encoder and kept official hyperparameters for a fair comparison. 
Most of these methods were only tested on large-scale datasets, such as the ImageNet~\cite{imagenet_cvpr09}, usually with very deep convolutional encoders. 
Thus, it naturally raises the question of how these methods would perform on smaller datasets, using faster and not-so-deep encoders, without enormous batch sizes and trained on modest GPUs.
Our experiments highlight some of the limitations of training these data-hungry methods on limited resources.

\subsubsection{Cross-domain transfer}

To evaluate how well the representations learned by CARL transfer between different distributions, the first two columns of Table~\ref{tab:results-cross-domain-and-semi-sup} show experimental results on cross-domain transfer.
We took a self-supervised encoder trained on the CIFAR-10 dataset and used it as a feature extractor to train a linear classifier (for 50 epochs) to solve a 100-class classification task using the CIFAR-100 dataset (and vice versa).
Representations learned by CARL outperform (on average) the main competitors in both setups.

\subsubsection{Semi-supervised learning}
The third and fourth columns of Table~\ref{tab:results-cross-domain-and-semi-sup} display results on semi-supervised learning.
First, we trained self-supervised models (for 200 epochs) on the unsupervised portion of the STL-10 dataset. 
Afterward, we fine-tuned the encoders (for 50 epochs) using different portions (\SI{1}{\percent} and \SI{10}{\percent}) of labeled data from the supervised set of STL-10.
Here, representations learned by CARL outperform competitors on the \SI{10}{\percent} setup and perform on pair with SimCLR on the more extreme case of only \SI{1}{\percent} of labeled data.

\subsubsection{Fine-tuning}
Lastly, similar to the semi-supervised evaluation, the fifth column of table~\ref{tab:results-cross-domain-and-semi-sup} shows results on fine-tuning the self-supervised encoder (trained on the unsupervised STL-10 for 200 epochs) using the entire supervised STL-10 dataset (for 50 epochs).
CARL's representations outperform competitors by at most 2.\SI{43}{\percent} average accuracy.

\begin{table*}
  \footnotesize
  \sisetup{
    round-mode = places,
    round-precision=2,
    table-text-alignment=center,
    table-format=2.2(2),
    separate-uncertainty,
    detect-all,
  }
  \setlength{\colsep}{.65em}
  \begin{center}
  \caption{Results are reported using mean accuracy (percentage) and standard deviation over three runs (except for DS ImageNet where we report a single run) on the linear evaluation protocol.}
  \label{tab:results-small-datasets}
  \vspace*{-10pt}
    \begin{tabular}{
        lSSSSSSS[table-format=2.2]S[table-format=2.2]
      }
      \toprule
      & \multicolumn{2}{c}{CIFAR-10} & \multicolumn{2}{c}{CIFAR-100} & \multicolumn{2}{c}{STL-10} & \multicolumn{2}{c}{DS ImageNet} \\
      \cmidrule(lr){2-3} \cmidrule(lr){4-5} \cmidrule(lr){6-7} \cmidrule(lr){8-9}
      Method & {100} & {200} & {100} & {200} & {100} & {200} & {100} & {200}\\
      \midrule
      Supervised   & 87.76 (007) & 89.40 (022) & 59.23 (028) & 60.59 (009) & 73.11 (024) & 74.76 (025) & 35.06 & 43.5 \\
      \midrule
      BYOL~\cite{grill2020bootstrap}   & 68.90 (024) & 76.46 (037) & 37.26 (047) & 45.68 (016) & 76.58 (078) & 79.64 (020) & 28.12 & 30.96 \\
      SimCLR~\cite{chen2020simple} & 72.64 (034) & 75.41 (047) & 41.64 (014) & 44.94 (011) & \bfseries 77.36 (039) & 80.57 (066) & \bfseries 28.20 & 30.32 \\
      MoCo~\cite{he2020momentum}   & 65.57 (065) & 71.62 (104) & 39.12 (043) & 44.35 (041) & 71.50 (096) & 75.46 (128) & 24.82 & 28.67\\
      PCL~\cite{li2020prototypical}   & 66.37 (023) & 71.29 (054) & 35.97 (053) & 40.1 (076) & 69.61 (093) & 70.56 (041) & 22.60 & 26.18\\
      SwAV~\cite{caron2020unsupervised}   & 72.75 (065) & 75.93 (048) & 39.98 (083) & 43.20 (035) & 70.89 (055) & 74.35 (047) & 18.389 & 25.313 \\
      CARL   & \bfseries 73.39 (031) & \bfseries 78.94 (052) & \bfseries 42.91 (026) & \bfseries 48.85 (079) & 76.9 (012) & \bfseries 81.95 (010) & 24.618 & \bfseries 31.856 \\
      \bottomrule
    \end{tabular}
  \end{center}
\end{table*}

\begin{table*}
  \footnotesize
  \sisetup{
    round-mode = places,
    round-precision=2,
    table-text-alignment=center,
    table-format=2.2(2),
    separate-uncertainty,
    detect-all,
  }
  \setlength{\colsep}{.65em}
  \begin{center}
  \caption{Results are reported using mean accuracy (percentage) and standard deviation over three runs. Best results in bold. Cross-Domain Transfer: We trained CARL on the unlabeled CIFAR-10 dataset and performed linear evaluation on the labeled CIFAR-100 (10 $\rightarrow$ 100), and vice versa (100 $\rightarrow$ 10). Fine-tune: We trained CARL on the unlabeled STL-10 dataset and fine-tune it on the labeled STL-10.}
  \label{tab:results-cross-domain-and-semi-sup}
  \vspace*{-10pt}
    \begin{tabular}{
        lSSSSS
      }
      \toprule
      & \multicolumn{2}{c}{Cross Domain Transfer} & \multicolumn{2}{c}{Semi-supervised} & \multicolumn{1}{c}{Fine-tuning}\\
      \cmidrule(lr){2-3} \cmidrule(lr){4-5} \cmidrule(lr){6-6}  %
      Method & {C-10 $\rightarrow$ C-100} & {C-100 $\rightarrow$ C-10} & {\SI{1}{\percent}} & {\SI{10}{\percent}} & {\SI{100}{\percent}}\\
      \midrule
      Supervised   & 38.11 (041) & 68.54 (042) & 52.97 (007) & 69.87 (003) & 76.11 (001)\\
      \midrule
      BYOL~\cite{grill2020bootstrap}   & 43.47 (020) & 67.23 (045) & 36.64 (043) & 65.7 (098) & 85.72 (010)\\
      SimCLR~\cite{chen2020simple} & 40.39 (022) & 67.4 (030) & \bfseries 53.66 (011) & 73.92 (026) & 86.97 (003)\\
      CARL   & \bfseries 43.47 (002) & \bfseries 67.89 (023) & 52.26 (114) & \bfseries 74.85 (033) & \bfseries 88.15 (037)\\
      \bottomrule
    \end{tabular}
  \end{center}
\end{table*}

\section{Conclusions}

In this work, we presented Consistent Assignment for Representations Learning (CARL). 
An unsupervised method that learns visual representations by forcing augmented versions of an image to be consistently assigned over a finite set of learnable prototypes.
Unlike contrastive learning methods, we propose a higher-level pretext task that operates over the distributions of views instead of directly optimizing the view's embeddings.
Our method differs from recent work that merges clustering with contrastive learning since it does not require pre-clusterization steps or non-differentiable algorithms to solve the cluster assignment problem. 
Instead, we learn a set of general prototypes that act as energy anchors for the views' representations, entirely online using gradient descent.
We studied some of the main components of CARL and the effects of different configurations of hyper-parameters. 
Lastly, our results show that representations learned by CARL surpass many of the recent contrastive learning and clustering methods without resorting to a large number of negative samples or extra encoders. 
The representations are qualitatively measured using many datasets and benchmarks, including linear evaluation, semi-supervised and finetuning.

\appendix
\section{Implementation Details}
\label{sec:appendix-A}
For evaluating CARL's performance against its competitors, all methods were trained for \num{200} epochs using the same cosine learning rate scheduler and the same batch size of 256 observations.
If not specified otherwise, we kept all the hyperparameters fixed as the original implementations.

\subsection{Datasets}

To properly validate our model's performance, we opted to use many standard computer vision datasets. 
For small to medium classes, we used the CIFAR-10/100 dataset family. 
Each dataset contains \num{60000}, $32 \times 32$ RGB images, with \num{50000} images for training and \num{10000} for testing. 
CIFAR-10 has \num{10} classes with \num{6000} images per class, and CIFAR-100 has \num{600} images for each of the \num{100} classes.

We used the STL-10 unsupervised and supervised sets for fine-tuning and semi-supervised experiments. 
The unsupervised portion contains \num{100000}, $96 \times 96$ RGB images, and the supervised set has \num{5000} images from \num{10} classes.

For datasets with a significant number of classes, we used the Tiny-ImageNet and a downsampled version of the original ImageNet dataset, referred to as DS ImageNet.
Tiny-ImageNet contains \num{100000} $64 \times 64$ RGB images balanced across \num{200} different categories. 
The downsampled version of ImageNet has \num{1000} classes with \num{1281167}, $64 \times 64$ RGB images.

\subsection{Backbones}

For all experiments, the encoder~$f$ comprises a ResNet-18 backbone followed by a non-linear \num{2}-hidden layer fully-connected network, as projection head.
For CARL, SwAV, and BYOL, the projection head also has a batch normalization layer. 
The function~$f$ encodes an image $x_i$ into a \num{128}-dim representation.
Specifically, the projection head receives a $512$-dim vector from the final average pooling layer of the ResNet-18 encoder.
The hidden layer of the projection head has $512$~neurons.
For all methods, the dimensionality of the embedding vector~$z_i$ and the complexity of both the encoder and projection head are equivalent.

\subsection{Augmentations}

To create views for optimizing CARL, we used the same pipeline of data augmentations proposed by~\citet{chen2020simple}.
First, we apply a random crop resize operation, which randomly extracts a portion of the image from 0.08 to 1.0 of the original size.
Second, we apply a horizontal flipping operation with a \SI{50}{\percent} probability. 
There is an \SI{80}{\percent} probability of jittering the pixels of the image, altering its brightness, contrast, saturation, and hue.
Then, a \SI{20}{\percent} chance of a grayscale conversion, a \SI{50}{\percent} chance of gaussian blurring, followed by normalization using the dataset's mean and STD.

\subsection{CARL}

CARL is trained using SGD with momentum and a cosine learning rate decay scheduler~\cite{loshchilov2016sgdr} (without restarts) starting at \num{0.6} and decaying to \num{0.0006}. 
We use a weight decay penalty of $5 \times 10^{-4}$ and the LARS optimizer~\cite{you2017large}, but using pure SGD produces similar results.
To choose the number of general-prototypes, we follow the findings from our hyper-parameter exploration, see Section~\ref{sec:ablations}. 
For CIFAR-10/100, and STL-10, we trained CARL using \num{100} prototypes for the first dataset and \num{300} for the last two. 
Following SwAV, for the downsampled ImageNet dataset, we used \num{3000} prototypes. 

\subsection{Other methods}

To compare the performance of CARL with other implementations, except for MoCo, we developed our versions of SimCLR, BYOL, SwAV, and PCL by adapting their original code repositories and performing the minimal changes necessary to meet our requirements. 
The core changes to the official implementations regard (1) adaptation to other datasets, (2) support for single GPU training, (3) change of backbone encoder (from ResNet-50 to ResNet-18), (4) when necessary, change hyperparameters to achieve better results or to comply with our computational budget.

\subsubsection{PCL}~\cite{li2020prototypical}. For training on CIFAR-10/100 and STL-10, the number of prototypes is linearly scaled based on the number of true classes. 
For CIFAR-10 and STL-10, we set the number of clusters to $\{250, 500, 1000\}$, and for CIFAR-100, we used $\{2500, 5000, 10000\}$. 

\subsubsection{SwAV}~\citet{caron2020unsupervised}. For CIFAR-10/100 and STL-10, we used the same number of prototypes used to train CARL\@. 
The number of prototypes was set to \num{100} for CIFAR-10, and \num{300} for both CIFAR-100 and STL-10. 
We used the standard configurations from their original repositories to train both PCL and SwAV on the downsampled version of the ImageNet dataset. 

\subsubsection{BYOL and SimCLR}
We empirically found that removing the LARS optimizer (from their official implementations) produced better results in our datasets.

We performed a simple grid search to find an optimal learning rate of \num{0.03}, which improved over original parameters.

Originally, BYOL~\cite{grill2020bootstrap} uses a denser representation vector of \num{256}-dim and a more complex projection head with a hidden layer containing \num{4096} neurons.
BYOL employs batch sizes of 4096 observations.
All methods use the same projection head architecture, and the extra BYOL's predictor head follows the same architecture as the projection head.

For SimCLR~\cite{chen2020simple}, we empirically found that a temperature parameter of 0.2, instead of 0.5 described in the paper~\cite{chen2020simple}, is best for our datasets. 
We could not reproduce the original training setups for SimCLR and BYOL due to limited hardware.
To ensure a fair comparison, we used equal batch sizes of \num{256} for all implementations. This allows SimCLR to produce \num{130560} pairs against \num{256} pairs used in CARL.

\subsubsection{Supervised} To establish an upper bound for performance comparisons, we trained a supervised ResNet-18 using SGD and a learning rate of 0.03 with a cosine learning rate decay (without restarts) that decreases to 0.
For data augmentation, we used a lighter version of the MoCo's~\cite{he2020momentum} augmentation pipeline composed of (1) random crop resize, (2) random horizontal flip (\SI{50}{\percent} chance), (3) random color jitter, (4) random gray scaling (\SI{20}{\percent} chance), and (5) random Gaussian blurring (\SI{20}{\percent} chance). 

\subsection{Computing infrastructure}

Experiments have been conducted on a workstation with 4 NVIDIA GeForce RTX 2080 Ti GPUs (11 GB of RAM). We adapted all methods to fit in a single GPU.

\begin{acks}
This study was financed in part by the Coordenação de Aperfeiçoamento de Pessoal de Nível Superior---Brasil (CAPES)---Finance Code 001.
\end{acks}

\bibliographystyle{ACM-Reference-Format}
\bibliography{abrv,sample-bibliography}

\end{document}